\documentclass{article} 
\usepackage{iclr2021_conference,times}

\usepackage{amsmath,amsfonts,bm}

\def\eqref#1{equation~\ref{#1}}

\def\1{\bm{1}}

\DeclareMathAlphabet{\mathsfit}{\encodingdefault}{\sfdefault}{m}{sl}
\SetMathAlphabet{\mathsfit}{bold}{\encodingdefault}{\sfdefault}{bx}{n}

\usepackage{hyperref}
\usepackage{url}
\usepackage{listings}
\usepackage[ruled,vlined]{algorithm2e}
\usepackage{graphicx}
\usepackage{subfigure}
\usepackage{booktabs}
\usepackage{multirow}

\title{Independent Density Estimation \\ For Compositional Generalization}

\author{
Jiahao Liu \\
Mobi.ai\\
\texttt{jiahao@takemobi.com} \\
\And
Senhao Cao \\
Orcava Inc.\\
\texttt{senhao.cao@orcava.ai} \\
}

\iclrfinalcopy 
\begin{document}

\maketitle

\begin{abstract}
Large-scale Vision-Language models have achieved remarkable results in various domains, such as image captioning and conditioned image generation. Nevertheless, these models still encounter difficulties in achieving human-like compositional generalization. In this study, we propose a new method called Independent Density Estimation (IDE) to tackle this challenge. IDE aims to learn the connection between individual words in a sentence and the corresponding features in an image, enabling compositional generalization. We build two models based on the philosophy of IDE. The first one utilizes fully disentangled visual representations as input, and the second leverages a Variational Auto-Encoder to obtain partially disentangled features from raw images. Additionally, we propose an entropy-based compositional inference method to combine predictions of each word in the sentence. Our models exhibit superior generalization to unseen compositions compared to current models when evaluated on various datasets.

\end{abstract}

\section{Introduction} \label{intro}
In recent years, large-scale Vision-Language models have achieved remarkable success across various domains, including image captioning, conditioned image generation, and more. These deep learning models operate by processing data from both visual and textual modalities and subsequently fusing the two information streams. A notable example is the CLIP model \cite{radford2021learning}, which jointly trains an image encoder and a text encoder to predict accurate pairings of a batch of $(image, text)$ training examples. More specifically, it uses a Vision Transformer (ViT) \cite{dosovitskiy2020image} to encode visual inputs and a Transformer-based \cite{vaswani2017attention} text encoder to encode input sentences. After that the image and text embeddings are projected to the same space and a cosine similarity is computed to represent whether this text input matches the visual input. This paradigm achieves huge success and is adopted in many areas, such as training an agent to follow language instructions in situated environments \cite{chevalier2018babyai}. 

Despite the significant accomplishments of Vision-Language models, they face challenges in achieving human-level compositional generalization \cite{lewis2022does}, a crucial goal that involves forming new concepts by combining existing ones. For example, after grasping the meanings of individual words or concepts like a \textit{blue cube} and a \textit{red sphere}, we can seamlessly understand \textit{blue sphere} by drawing upon our knowledge of the components \textit{blue} and \textit{sphere}, even when encountering this combination for the first time. 

In this work, we focus on a simple question: given a sentence description of an object, such as \textit{large blue cube}, can deep learning models predict the corresponding image even if this combination is unseen in the training stage?  We propose a new method named Independent Density Estimation (IDE) to solve this problem. Unlike the previous paradigm which uses a single embedding to represent the whole input sentence and learn it's correlation with the visual inputs, we seek to learn which words in the input sentence describe which parts of the visual inputs. After matching words in the sentence with the corresponding features in the image, we utilize an entropy-based method to compose words in sentences to generate new predictions.

Our paradigm is built on top of disentangled visual representations \cite{higgins2018towards}, which is a widely explored representation formulation that emphasizes explicit compositionality. It posits that a change in one factor of the observation corresponds to a change in a single factor of the representation. We develop two models based on the philosophy of IDE. The first model takes a fully disentangled visual representation as input, and learns how each word is associated with each feature channel. Next, we demonstrate that we can leverage Variational Auto-Encoders \cite{kingma2013auto} to obtain partially disentangled features from raw images, which also enables us to learn meaningful relationships between the visual representation and language descriptions. 

\begin{figure}[htbp]
\centering
\includegraphics[width=0.9\linewidth]{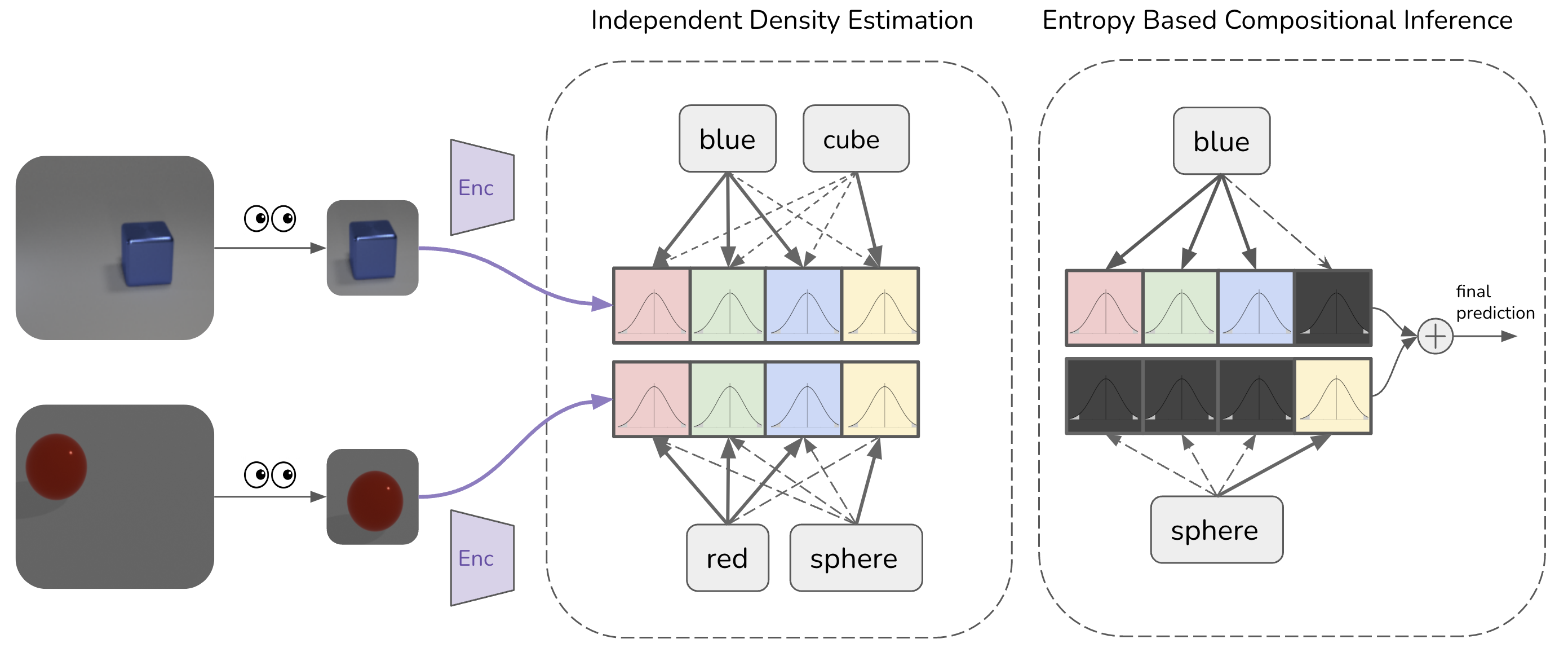} 
\caption{
Demonstration of how our method generalizes from \textit{blue cube} and \textit{red sphere} to unseen combinations such as \textit{blue sphere}. \textit{(left)} We utilize a mechanism similar to the human eye's object detection process to locate the object in the image, then feed the corresponding image patch into a VAE encoder to extract the disentangled latent representation. \textit{(middle)} We train density estimation networks to predict the value of each dimension in the representation for each word, and learn the certainty about the predictions. Bold arrows mean higher certainty and dashed ones are lower. \textit{(right)} During inference, we make independent predictions for each individual word in the description and compose them based on their respective prediction certainties. The black dimensions refer to predictions with lower certainty, which have less impact on the final prediction.
}
\label{fig:model}
\end{figure}

The main contributions of this paper include the following:

\begin{enumerate}
\item The paper presents Independent Density Estimation, a new method to learn connections between words with visual features that enables compositional generalization.
\item We proposed an entropy-based compositional inference method to compose the prediction from each word in a sentence.
\item We present a model that takes fully disentangled representations as inputs and demonstrate how this approach can be extended to handle raw visual inputs.
\item We evaluated our models on various datasets and showed that they can better generalize to unseen compositions compared with current models. 
\end{enumerate}

\section{Background}
\label{background}

\textbf{Object Selection Task}  \quad Our primary objective is to develop machine learning models capable of accurately selecting the corresponding image when provided with a textual description of an object, such as \textit{"a blue sphere"}. Formally, let an image be denoted as $x \in X$, and let a textual description be defined as a sequence of words $w = [w_0, \dots, w_l]$ with length $l$. Here, $X$ and $W$ denote the image and language description spaces, respectively. Given a set of candidates $ D = \{ x_0 \dots x_m \}$ and a language description $w$, the object selection task requires the model to predict the \textbf{unique} correct image $x_i$ for $i \in [0,m]$. We aim to learn a score function $S:X \times L \rightarrow \mathbb{R}$ which represents the compatibility between a description $w$ and an image $x$.  Given $w$ and a set of image candidates $D = \{ x_0 \dots x_m \}$, the selected image is the one with the highest score, i.e., $argmax_i(S(x_i,w))$ .

\textbf{Compositional Generalization}  \quad We specifically focus on the compositional generalization aspect of this problem, that is, given a description of an object, such as "blue sphere," can deep learning models identify the corresponding image even if this particular combination was not encountered during the training stage? To this end, we consider a limited setting: only a single object is allowed to appear in the image, and we restrict the language description to $w = [w_0, \dots w_l]$ where each word $w_i$ describes one attribute of the object present in the image, such as shape, size, or color.

\textbf{Vision-Language Model Paradigm} \quad A considerable amount of research has been dedicated to matching language descriptions with their corresponding images. Most of these studies, including CLIP, follow a similar approach. To process image information, researchers use a pre-trained visual encoder to obtain visual embeddings. Popular choices are ResNet \cite{he2016deep} and ViT \cite{dosovitskiy2020image}, which exhibit strong performance on visual tasks like image segmentation \cite{chen2021transunet} and video prediction \cite{weissenborn2019scaling} and are considered effective at extracting meaningful representations from raw image inputs. To handle text descriptions, researchers commonly apply a Transformer-based architecture to encode input sentences, capturing the underlying linguistic structure and relationships between words. After obtaining the image and sentence embeddings, researchers map them to the same feature space using simple fully-connected layers. Models are trained using Contrastive Learning framework \cite{le2020contrastive}: the distance between the embeddings of image-text pairs is minimized if they match or maximized if they don’t.

Despite their success, models following the aforementioned paradigm often struggle to achieve compositional generalization. In the following section, we discuss how to improve models' compositionality through Independent Density Estimation and entropy-based inference method.

\section{Method}
\label{method}

\subsection{Independent Density Estimation} \label{method:ide}

In order to learn the relationship between language and visual inputs, finding effective representations for the images is crucial since they exist in a high-dimensional space. In contrast to previous methods that rely on visual encoders pre-trained on classification tasks, we aim to use disentangled representations. Disentanglement involves separating the generative factors of observations into different factors of low-dimensional representations, such that each dimension or channel of the representation has a distinct semantic meaning that can be related to words in the language description. This section assumes that we have obtained a fully disentangled representations of the visual inputs and explains how we can use the Independent Density Estimation framework to learn the relationship between the language inputs and these representations. Section \ref{method:dis} discusses how to learn such representations.

Given an visual input $x$, the disentangled representation is a vector $f \in \mathbb{R}^d$ where $d$ is the number of the vector dimensions. Take the middle part of Figure \ref{fig:model} as an example: a disentangled representation is a four dimensional vector $[r,g,b,s] \in \mathbb{R}^4$, where the first three dimensions encode the color information in $(r,g,b)$ format, and the last one dimension corresponds to shape. The corresponding language description $w$ is a list of word $[w_0, \dots w_l]$. Now we want to learn the connection between description $w$ and feature $f$.  Given a word description \textit{red}, the first three dimension should be centered arround $(1, 0, 0)$, while nothing can be inferred for the fourth dimension. On the contrary, if we \textbf{only} observe the word \textit{cube}, we can predict the value in fourth dimension while having no idea about the first three channels. Therefore it is possible to build connection between word tokens and specific feature dimensions: if word $w_i$ is related to a certain dimension $j$ in the disentangled representation, the value of $f_j$ would be predictable after the observation of word $w_i$. Note that processing each word in the description \textbf{independently} is important in this situation. For example, given visual input and language description pair, represented as $( [1,0,0,1], [\text{\textit{red}, \textit{cube}}])$, if we use a standard transformer to encode the text input and obtain the embedding for the word \textit{red}, it will aggregate contextual  information from the word  \textit{cube}, and then the fourth dimension becomes predictable. Then we lose the ability to identity which word is related to which dimension.

We assume that conditioned on the word $w_i$, the value of $f_j$ follows a Gaussian distribution, i.e, $P(f_j|w_i) = \mathit{N}\left(\mu_{ij}, \sigma_{ij}\right)$ .  We build a density estimation network \cite{magdon1998neural} to estimate the distribution. More concretely, each density estimation network takes in a word embedding and use $k$ fully connected layers to estimate the distribution parameter $\mu_{ij}$ and $\sigma_{ij}$. We train the density estimation network through maximize the likelihood of training examples, and the corresponding loss function is Gaussian negative log likelihood loss:
\begin{equation}
    \mathcal L = \sum_{ij}\frac{1}{2} \left[\log\ \hat\sigma_{ij}^2 + {\frac {(f_{ij}-\hat\mu)^{2}}{\hat\sigma_{ij}^{2}}}\right]
\end{equation}
where $f_{ij}$ is the input visual representation and $\hat\mu_{ij}$ and $\hat\sigma_{ij}$ is the network estimations.

\subsection{Entropy Based Compositional Inference} \label{method:epy}
In this section, we discuss how to build a score function $S : X \times L \rightarrow \mathbb{R}$ that measures the compatibility between an image $x \in X$ and textual description $w = [w_0 \dots w_l] \in L$. This score can be used to solve \textbf{object selection task}: given $w$ and a set of image candidates $D = \{ x_0 \dots x_m \}$, the selected image is the one with the highest score, i.e., $\text{prediction} = argmax_i(S(x_i,w))$. The high level idea is to infer a "prototype" representation $\hat f$ from the given prompt $w$. We take the L2 distance between inferred representation $\hat f$ and actual representation $f$ as the score value, i.e., $S(w, x)= \|f,\hat f\|_2$. 

To infer a "prototype" representation $\hat f$ from the textual description $w$, we need to combine the knowledge in each individual density estimation network together to form the prediction. After learning the Gaussian distribution parameter $u_{ij}$ and $\sigma_{ij}$, we aim at combine these predictions from each word $w_i$ in the description $w$ to form a final prediction $\hat f \in \mathbb{R}^d$ where $d$ refers to the dimension of input representations. The reason why we formalize the learning task in the last section as density estimation instead of regression is because it allows us to quantify the uncertainty of prediction through compute the entropy of Gaussian distribution using the following equation:

\begin{equation}
    \begin{aligned}
    e_{ij}  &=-\mathbb{E}\left[\log \mathcal{N}\left(\mu_{ij}, \sigma_{ij}^2\right)\right] \\
    & \stackrel{ }{=} \frac{1}{2} \log \left(2 \pi \sigma_{ij}^2\right)+\frac{1}{2} .\end{aligned}
\end{equation}

Lower entropy value means higher certainty about the prediction of $\hat f_j$ after observing $w_i$, and vice versa. Then we can computer the \textbf{information gain} $g_{ij} = E - e_{ij}$ and use it as the weight to combine the prediction. Here $E$ is a hyper-parameter representing the entropy before observation. Since the entropy of continuous distribution can be infinitely large and we only care about the relative gain of information, in practice we can choose any reasonable large value to make sure $g_{ij}$ is positive.  Given a language description $w=[w_0, \dots w_l]$, we can predict a representation vector $\hat f \in R^d$, where each dimension $\hat f_j$ is the aggregation of predictions of all words weighted by $g_{ij}$: 
\begin{equation}
    \hat f_j = \sum_i\frac{exp (g_{ij} / \tau)}{\sum_i exp(g_{ij} / \tau ) } \cdot u_{ij}
\end{equation}
Here $\tau$ is the temperature parameter used to control the inference bias. As show in the right part of Figure \ref{fig:model}, this inference step naturally supports compositional generalization, as the each individual word's relationship with visual feature is figured out during the training stage. 

\subsection{Learning Disentangled Representation For IDE} \label{method:dis}
State-of-the-art unsupervised disentanglement models are primarily built on variational generative models, such as Variational Auto-Encoder (VAE). However, obtaining fully disentangled visual representations from raw images remains a challenge due to the complexity of the generative process. A major challenge is disentangling an object's position from its visual appearance. \cite{watters2019spatial} addressed this issue by broadcasting the latent vector across the space, but this approach requires objects to appear in all different positions during training. As shown in the left part of Figure \ref{fig:model}, we propose an alternative method. Instead of learning a disentangled representation from the whole image, we first identify the object's position and then clip the image patch centered around the object. Then we train a VAE to extract disentangled representation of that image patch.

To circumvent the need for training a neural network for object segmentation or detection, we adopt the concept of V1 saliency maps from neuroscience to determine the object's location within the image. The V1 saliency map assigns a saliency value to each pixel based on the likelihood of it being the location of an object. This map guides the eye-movement process so that human can easily detect objects in the scene. Saliency values are computed based on the iso-feature suppression principle: nearby neurons with the same type of activation suppress each other, leading to more salient features exhibiting higher activation. Consider an input image $x\in \mathbb{R}^{H \times W \times C}$ with height $H$, width $W$ and color channel $C$. The saliency value at each point is:

\begin{equation}
    s_{ij} = B - \sum_{x'_{ij}\in nei(x_{ij})}(1- \|x_{ij} - x'_{ij}\|_2)
\end{equation}

Here $B$ is a hyper-parameter which represents the base activation rate of the saliency map, and $x'_{ij}$ is the neighbor pixel of $x_{ij}$. After obtain the saliency map of an image, we perform a search algorithm on the map to find the bounding box of the object. 

The clipped image output can be fed into a standard VAE, and partially disentangled representations (also known as latents) can be extracted from this image patch. Subsequently, we follow the steps mentioned in previous sections to learn the relationship between language descriptions and visual representations.

\section{Experiment}

\subsection{Object Selection using Disentangled Representations}
In this section, we evaluate our method described in Section \ref{method:ide} and \ref{method:epy} on fully disentangled representations. We use three datasets, one is a toy dataset crafted by ourselves and the other two are obtained from reinforcement learning environments. Experiments shows that our method can better generalize to unseen compositions under different settings.

\subsubsection{Dataset and Baseline}

\textbf{Toy Dataset} \quad We manually create a toy dataset to probe our model's performance. We use a five dimensional vector to represent the attribute of an object. The first three dimensions encode color information in \textit{(red,green,blue)} format. The fourth dimension represents different materials, including \textit{rubber} and \textit{metal}. The fifth represents shapes, including \textit{ball}, \textit{triangle}, and \textit{rectangle}. For each dimension, we add a small Gaussian noise with $\mu=0$ and $\sigma=0.05$. We enumerate all the possible combinations to generate 9000 \textbf{(representation vector, language description)} pairs as our dataset. 

\textbf{BabyAI and AI2Thor Dataset} \quad Object-oriented tasks are common in reinforcement learning (RL) settings. For example, an agent is given the prompt "go to the blue ball" and an important step is to identify the target object from the environment. Furthermore, the representation used in such RL environments are usually fully disentangled, which provides us a good environment for evaluating our model. We focus on two RL environments: BabyAI\cite{chevalier2018babyai} and AI2Thor \cite{kolve2017ai2}, which are widely used benchmarks for testing grounded language learning. We obtain \textbf{(representation vector, language description)} pairs from the RL environments. In BabyAI, each object is encoded using 3 integer values: one describing the shape of object contained in the cell, one describing its color, and a state indicating whether doors are open, closed or locked.  All the three dimensions are encoded in a symbolic way, for example, the color dimension is encoded as: \verb|{"red": 0, "green": 1, "blue": 2, "purple": 3, "yellow": 4}|. We sample objects in BabyAI environments and generate corresponding descriptions, such as \textit{"a blue door"} or \textit{"a red ball"} for each object representation. AI2Thor use a six dimensional vector and each dimension has a unique semantic meaning respectively: [\textit{mass}, \textit{temperature}, \textit{if toggled}, \textit{if broken}, \textit{if dirty}, \textit{object shape}]. Unlike other dimensions, the first two dimensions are represented by floating-point values normalized between 0 and 1. We describe \textit{mass} values smaller than 0.4 as \textit{light} and larger ones as \textit{heavy}. Similarly, for \textit{temperature} values lower than 0.3, we describe it as \textit{cold}, between 0.3 and 0.6, we describe it as \textit{room-temperature}, and between 0.6 and 1, we describe it as \textit{hot}. 

\textbf{Leave-out Set} \quad In order to test models' compositional generalization ability, we leave out part of the dataset whose descriptions have not appeared in the training stage. 
For our toy dataset, we leave out the following combinations: \textit{"red metal ball", "red rubber triangle", "blue rubber rectangle", "blue metal triangle", "green metal triangle", "green rubber ball"}. For BabyAI dataset, we leave out \textit{"green locked door", "blue locked door", "red locked door", "blue ball", "green ball", "yellow key", "grey wall", "blue box", "grey floor"}. For AI2Thor dataset, we leave out \textit{"a cold apple", "a unbroken countertop", "a hot bottle", "a light creditcard", "a dirty bowl"}.

\textbf{Baseline} \quad Following the spirit of CLIP, we build a toy-CLIP model that works on disentangled feature input. We use two fully-connected layers to encode the input disentangled representation. We use Leaky-Relu activation function with slope equals to $0.1$. We use a standard LSTM to encode the language description. Similar to CLIP, the output from these two modals are mapped to a 16-dimensional vector space and cosine similarity is used to represent the compatibility of the input representation and description.

\textbf{Implementation of IDE} \quad We follow the description in Section \ref{method:ide} to implement our model. The training examples are \textbf{(representation vector, language description)} pairs. We randomly initialize a word vector of four dimensions to represent each word. Then we build a standard density estimation network with two fully-connected layers to estimate the Gaussian distribution parameter $\mu$ and $\sigma$. Then we follow the description in Section \ref{method:epy} compose the output of density network to form the final prediction $ \hat f \in \mathbb{R}^d$ where $d$ is the dimension of input representations. 

\subsubsection{Results on BabyAI and AI2Thor Dataset}  \label{expr:rl}

\begin{figure}[htbp]
\centering
\includegraphics[width=0.6\linewidth]{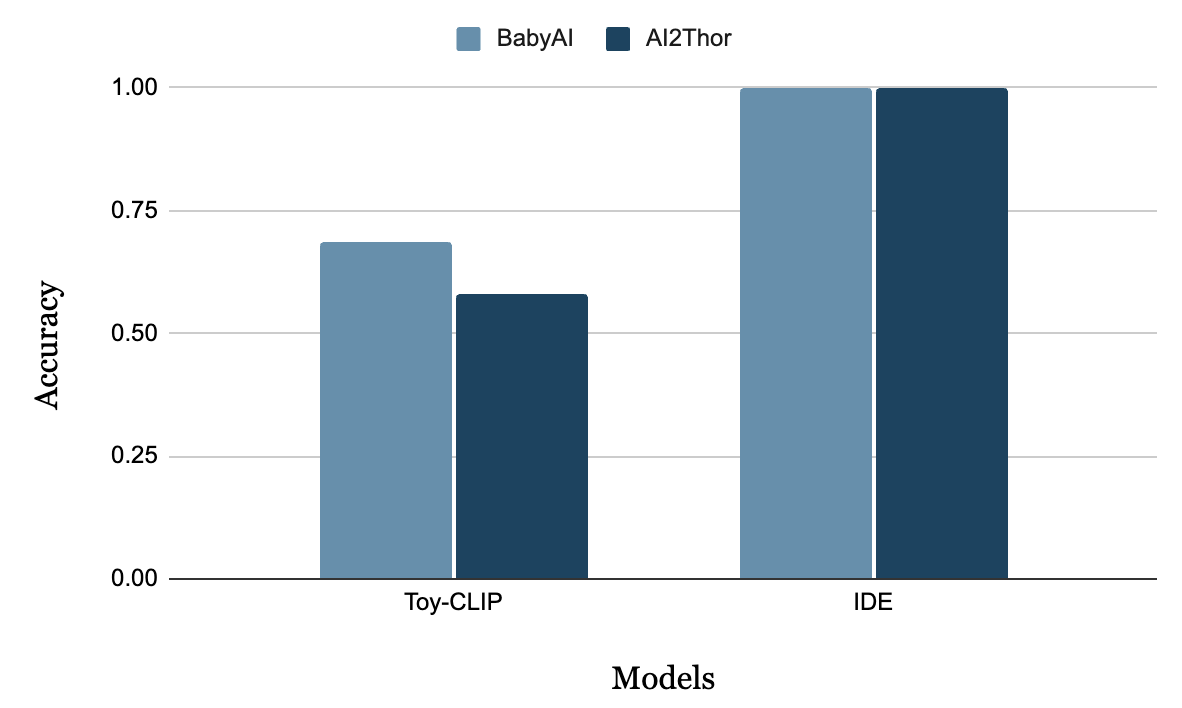} 
\caption{ Test accuracy on BabyAI and AI2Thor dataset.}
\label{fig:baby}
\end{figure}

As described in Section \ref{background}, the object selection task requires the model to predict the correct image from a set of candidates $ D = \{ x_0 \dots x_m \}$ based on description $w$. We construct a test example $(D', w')$ as the following: (1) the language description $w'$ is randomly sampled from the leave-out prompts, (2) $D' = \{x_0, \dots x_m \}$ where each image $x_i$ has a different language description in the leave-out set. This this experiment each image $x_i$ is substituted by a disentangled representation vector $f_i \in \mathbb{R}^d$. The total test size of each dataset is 3000.\\

Test result on BabyAI and AI2Thor dataset is shown in Figure \ref{fig:baby}. Our model (IDE) get 100\% accuracy in this setting. Toy-CLIP model get 68.8\% and 57.9\% on BabyAI and AI2Thor respectively. It demonstrates that our model can better generalize to unseen combinations.

\subsubsection{Results on Toy Dataset} \label{expr:toy}

\begin{figure}[htbp]
\centering
\subfigure[]{
\includegraphics[width=0.45\linewidth]{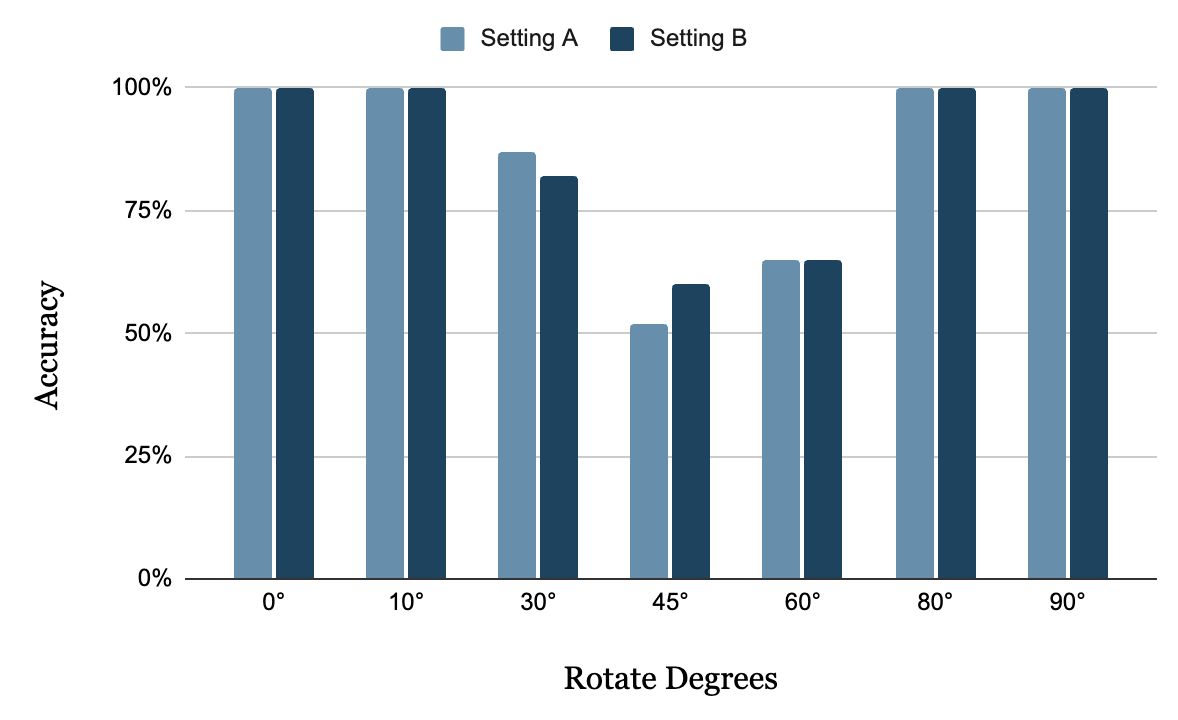}
\label{fig:rotate}
}
\subfigure[]{
\includegraphics[width=0.45\linewidth]{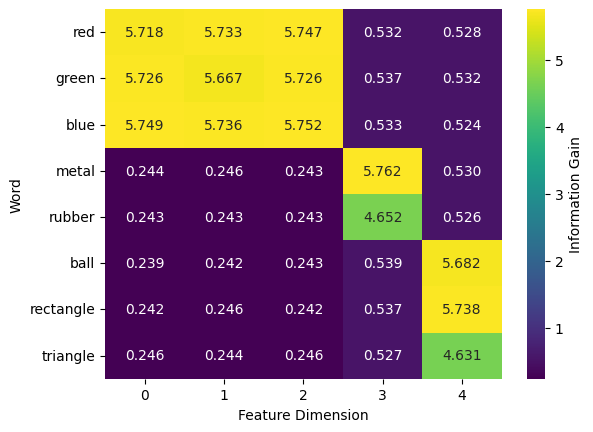}
\label{fig:entropy}
}
\caption{(a) shows performance of our model under different settings. (b) is a heatmap of information gain of each feature dimension conditioned on each word after training.}
\label{fig:both}
\end{figure}

Following the description in Section \ref{expr:rl}, we also construct test examples based on leave-out dataset. Additionally, we perform two different controls on our dataset:
\begin{enumerate}

\item \textbf{Rotate Input Representations} \quad For each input disentangled representation, we rotate the representation vector by $\theta$ degrees to manually make it entangled. Since our model is based on the assumption that the input representation is disentangled, by rotating the feature will lower the model's performance. In the experiment, we use $\theta \in \{0^{\circ}, 10^{\circ}, 30^{\circ}, 45^{\circ}, 60^{\circ}, 80^{\circ}, 90^{\circ}\}$.

\item \textbf{Partial Language Description} \quad In our daily life the description of an object often covers part of the object attributes. To study whether partial descriptions will influence our model's ability of compositional generalization, we manually drop attributes in the description during the training stage. Concretely, we randomly choose one of the first two attributes in the description and delete it. For example, we delete \textit{"rubber"} from \textit{"blue rubber triangle"} to get \textit{"blue triangle"}. In Figure \ref{fig:rotate}, we use \textit{setting A} to denote models trained on full descriptions and \textit{setting B} to refer models trained on partial descriptions.

\end{enumerate}

Figure \ref{fig:rotate} shows the model's performance on these two controlled settings. As we expected, when we increase the rotation angle from $0^{\circ}$ to $45^{\circ}$, the performance start to decline as the input representations start to entangle. When we increase the rotation angle from $45^{\circ}$ to $90^{\circ}$, the performance starts to improve because the representations become more disentangled again. The accuracy under setting A and B are similar to each other, showing our model is incentive to partial descriptions. Figure \ref{fig:entropy} demonstrate a heatmap of information gain $g_{ij}$ defined in Section \ref{method:epy}. We set the base entropy $E$ to $3.0$ to make sure every the $g_{ij}$ is positive. We can see clear connections between each word and feature dimensions. For example, material information such as \textit{metal} and \textit{rubber} are encoded in the fourth dimension.

\subsection{Object Selection using Raw Images}

\textbf{Blender dataset} \quad We manually create a dataset with a single object in a gray background (shown in Figure \ref{fig:model}). Similar to CLEVR \cite{johnson2017clevr}, we use Blender to render the image. There are four variables in the generative process: shape, size, color and position. The color feature contains \textit{red}, \textit{blue} and \textit{green}. The shape feature contains \textit{sphere}, \textit{cube} and \textit{cylinder}. For each generated image $x$, our language description $w$ is \textit{"[color] [shape]"}.  50 different images is generated for each description. We leave out all the \textit{"red cube", "blue cylinder"} and \textit{"green sphere"} for test and use the remaining to train our model. 

\textbf{Baseline} \quad We use standard CLIP implemented by OpenAI as our baseline. CLIP is a standard Vision-Language model trained on 400 million (image, text) pairs collected from the internet. CLIP applies a simple pre-training task of predicting which caption goes with which image, enabling zero-shot transfer of the model to over 30 downstream tasks.

\textbf{Implementation of IDE} \quad We follow the description in Section \ref{method:dis} to implement our model. We first train a VAE using the images in the training dataset. We use two convolutional layers to encode the input image, and then a fully connected layer to project the encoding to a four-dimensional latent space. Two deconvolutional layers is used to render reconstructed images from latent. We train VAE with batch size equals to 200 and learning rate equals to 0.001. After training VAE, we extract the latent representation of each image and follow the IDE framework to do estimation, and use entropy-based method to do inference.

\begin{table}[htbp]
\centering
\caption{Experiment Results and Meta Information of Models}
\label{tab:clip}
\begin{tabular}{@{}ccccc@{}}
\toprule
Model & Training set & Leave-out set & Model Parameters & Training Data Volume \\ \midrule
CLIP & 0.94 & 0.97 & 63M & 400M \\
Ours & 1.00 & 0.99 & 0.3M & 300 \\ \bottomrule
\end{tabular}
\end{table}

\textbf{Result} \quad Accuracy on training and testing dataset is shown in Table \ref{tab:clip}. Our model out performs CLIP model by 6\% on training set and 2\% leave-out set respectively. \cite{lewis2022does} reports that  after fine-tuning on CLEVR dataset \cite{johnson2017clevr}, CLIP's performance on leave-out set drops from 92.39\% to 78.54\%. This drop is because the model overfits the training set. We don't have the computational resource to fine tune the whole CLIP model, while we expect similar results on our Blender dataset, since it has almost the same data distribution as CLEVR. Notably, our model has a smaller parameter size compared to CLIP. Also, CLIP is pre-trained on large scale dataset, while ours is trained on only 300 images. This shows our model is not only better at compositional generalization, but also more parameter and data efficient.

\section{Related Work}

\textbf{Vision-Language Models} \quad Recent works such as CLIP \cite{radford2021learning}, CLOOB \cite{furst2022cloob}, and DeCLIP \cite{li2021supervision} bridge the vision and language modalities by learning a text encoder and an image encoder jointly with a contrastive loss, using large datasets consisting of (image, caption) pairs. CLIP learns expressive image embeddings directly from raw text, thereby leverages a much richer source of supervision than just labels. CLOOB equipped CLIP with  modern Hopfield networks to amplify co-occurrences and covariance structures of the original data. DECLIP demonstrated that they can learn generic visual features more efficiently by carefully utilizing the widespread supervision among the image-text pairs.
Recently \citet{lewis2022does} demonstrated that CLIP-based methods can not generalize to unseen compositions. We hypothesis that this is because these methods compress all the information in the whole sentence into a single embedding, losing the ability to identify which word describes which part of the visual feature. On the contrary, our method learns the correlation between each individual word and the image embedding, which demonstrated strong ability of compositional generalization.

\textbf{Compositional Generalization} \quad Previous research aimed at improving the compositional generalization of visual reasoning systems can generally be divided into two categories. The first group \cite{han2019visual} \cite{mao2019neuro} jointly learns concepts and metaconcepts representations from images and associated question-answer pairs, and apply neuro-symbolic reasoning modules to make inferences based on latent representations. The second group explicitly induce compositional inductive biases in to neural networks, such as visual grammars \cite{zhu2007stochastic} \cite{chen2007rapid}, compositional embeddings \cite{misra2017red} and neural module networks \cite{kuo2020compositional}. Unlike previous models, we propose to explicitly build connections between each word and a disentangled visual representation, and then compose the learnt connections in the inference stage.

\section{Conclusion}
In this work, we try to address the problem of compositional generalization in vision-language tasks. We propose a framework called Independent Density Estimation (IDE) to model the compatibility between the input and the description. Our model learns the connection between words and disentangled visual representations and composes these connections during the inference stage based on the entropy.

Our experiments on object selection tasks demonstrate that our model is able to generalize to unseen compositions and achieve better performance than the baselines. In addition, our model is more parameter and data efficient compared to existing methods such as CLIP.

We believe our work presents a promising direction in improving compositional generalization of vision-language models. Future work could extend our framework to other vision-language tasks and explore more complex language descriptions or representations.

\bibliography{iclr2021_conference}
\bibliographystyle{iclr2021_conference}

\end{document}